\documentclass[11pt]{article}
\usepackage[utf8]{inputenc}
\usepackage{acl2016}
\usepackage{times}
\usepackage{latexsym}
\usepackage{amsfonts}
\usepackage{url}
\usepackage{adjustbox}

\aclfinalcopy 

\title{Joint Learning of Sentence Embeddings for Relevance and Entailment}

\author{Petr Baudi\v{s},
	Silvestr Stanko \and
	Jan \v{S}ediv\'{y} \\
	FEE CTU Prague\\
	Department of Cybernetics\\
	Technick\'{a} 2, Prague,\\Czech Republic\\
	{\tt baudipet@fel.cvut.cz}}

\date{}

\begin{document}

\maketitle

\begin{abstract}
	We consider the problem of Recognizing Textual Entailment
	within an Information Retrieval context, where we must simultaneously
	determine the relevancy as well as degree of entailment for individual
	pieces of evidence to determine a yes/no answer to a binary
	natural language question.

	We compare several variants of neural networks for sentence embeddings
	in a setting of decision-making based on evidence of varying relevance.
	We propose a basic model to integrate evidence for entailment,
	show that joint training of the sentence embeddings to model
	relevance and entailment is feasible even with no explicit per-evidence
	supervision, and show the importance of evaluating strong baselines.
	We also demonstrate the benefit of carrying over text comprehension model
	trained on an unrelated task for our small datasets.

	Our research is motivated primarily by a new open dataset we introduce,
	consisting of binary questions and news-based evidence snippets.
	We also apply the proposed relevance-entailment model on a similar task
	of ranking multiple-choice test answers, evaluating it on
	a preliminary dataset of school test questions as well as
	the standard MCTest dataset, where we improve the neural model state-of-art.
\end{abstract}

\section{Introduction}

Let us consider the goal of building machine reasoning systems based
on knowledge from fulltext data like encyclopedic articles, scientific
papers or news articles.
Such machine reasoning systems, like humans researching a problem,
must be able to recover evidence from large amounts of retrieved
but mostly irrelevant information and judge the evidence to decide
the answer to the question at hand.

A typical approach, used implicitly in information retrieval
(and its extensions, like IR-based Question Answering systems \cite{YodaQAPoster2015}),
is to determine evidence relevancy by a keyword overlap feature (like tf-idf or BM-25 \cite{BM25})
and prune the evidence by the relevancy score.
On the other hand, textual entailment systems that seek to confirm hypotheses
based on evidence \cite{RTE1} \cite{SICK2014} \cite{SNLI}
are typically provided with only a single piece of evidence
or only evidence pre-determined as relevant, and are often restricted
to short and simple sentences without open-domain named entity occurences.
In this work, we seek to fuse information retrieval and textual entaiment
recognition by defining the \textbf{Hypothesis Evaluation} task
as deciding the truth value of a hypothesis by integrating numerous
pieces of evidence, not all of it equally relevant.

As a specific instance, we introduce the
\textbf{Argus Yes/No Question Answering} task.
The problem is, given a real-world event binary question like
\textit{Did Donald Trump announce he is running for president?}\ 
and numerous retrieved news article fragments as evidence,
to determine the answer for the question.
Our research is motivated by the Argus automatic reporting system for
the Augur prediction market platform. \cite{argus}
Therefore, we consider the question answering task
within the constraints of a practical scenario that has limited available dataset
and only minimum supervision.  Hence, authentic news
sentences are the evidence (with noise like segmentation errors, irrelevant participial phrases, etc.),
and whereas we have gold standard for
the correct answers, the model must do without explicit supervision on
which individual evidence snippets are relevant and what do they entail.

To this end, we introduce an open dataset of questions and newspaper evidence,
and a neural model within the Sentence Pair Scoring framework \cite{sps}
that (A) learns sentence embeddings for the question and evidence,
(B) the embeddings represent both relevance and entailment characteristics
as linear classifier inputs, and (C) the model aggregates all available evidence
to produce a binary signal as the answer, which is the only training supervision.

We also evaluate our model on
a related task that concerns ranking answers of multiple-choice questions given
a set of evidencing sentences.
We consider the MCTest dataset and the AI2-8grade/CK12 dataset that we introduce
below.

The paper is structured as follows.  In Sec.~\ref{sec:argus}, we formally outline
the Argus question answering task, describe the question-evidence dataset,
and describe the multiple-choice questions task and datasets.
In Sec.~\ref{sec:relwork}, we briefly survey the related work on similar problems,
whereas in Sec.~\ref{sec:model} we propose our neural models for joint
learning of sentence relevance and entailment.  We present the results in Sec.~\ref{sec:res}
and conclude with a summary, model usage recommendations and future work directions in Sec.~\ref{sec:concl}.

\section{The Hypothesis Evaluation Task}
\label{sec:argus}

Formally, the Hypothesis Evaluation task is to build a function $y_i = f_h(H_i)$,
where $y_i \in [0,1]$ is a binary label (\textit{no} towards \textit{yes})
and $H_i = (q_i, E_i)$ is a hypothesis instance in the form of question text $q_i$
and a set of $E_i = \{e_{ij}\}$ evidence texts $e_{ij}$ as extracted
from an evidence-carrying corpus.

\subsection{Argus Dataset}

Our main aim is to propose a solution to the Argus Task, where
the \textbf{Argus} system \cite{arguswp} \cite{argus}
is to automatically analyze and answer questions
in the context of the \textbf{Augur} prediction market platform.%
\footnote{\url{https://augur.net/}}
In a prediction market, users pose questions about future events
whereas others bet on the \textit{yes} or \textit{no} answer,
with the assumption that the bet price reflects the real probability
of the event.  At a specified moment (e.g.\ after the date of a to-be-predicted sports match), the
correct answer is retroactively determined and the bets are paid off.
At a larger volume of questions, determining the bet results may
present a significant overhead for running of the market.
This motivates the Argus system, which should partially automate
this determination --- deciding questions related to recent events
based on open news sources.

To train a machine learning model for the $f_h$ function,
we have created a dataset of questions with gold labels, and
produced sets of evidence texts from a variety of news paper using
a pre-existing IR (information retrieval) component of the Argus system.
We release this dataset openly.%
\footnote{\url{https://github.com/brmson/dataset-sts} directory data/hypev/argus}

To pose a reproducible task for the IR component, the time domain
of questions was restricted from September 1, 2014 to September 1, 2015,
and topic domain was focused to politics, sports and the stock market.
To build the question dataset, we have used several sources:
\begin{itemize}
	\item We asked Amazon Mechanical Turk users to pose questions, together with a golden label and a news article reference.
		This seeded the dataset with initial, somewhat redundant 250 questions.
	\item We manually extended this dataset by derived questions with reversed polarity (to obtain an opposite answer).
	\item We extended the data with questions autogenerated from 26 templates, pertaining top sporting event winners and US senate or gubernatorial elections.
\end{itemize}

To build the evidence dataset, we used the \textbf{Syphon} preprocessing component
\cite{argus}
of the Argus implementation%
\footnote{\url{https://github.com/AugurProject/argus}}
to identify semantic roles of all question tokens and produce
the search keywords if a role was assigned to each token.
We then used the IR component to query a corpus of newspaper
articles, and kept sentences that contained at least 2/3 of all
the keywords.
Our corpus of articles contained articles from The Guardian (all articles) and from the New York Times (Sports, Politics and Business sections).  Furthermore, we scraped partial archive.org historical data out of 35 RSS feeds from CNN, Reuters, BBC International, CBS News, ABC News, c|net, Financial Times, Skynews and the Washington Post.

For the final dataset, we kept only questions where at least
a single evidence was found (i.e.\ we successfuly assigned a role
to each token, found some news stories and found at least one
sentence with 2/3 of question keywords within).  The final size
of the dataset is outlined in Fig.~\ref{tab:dataset} and some
examples are shown in Fig.~\ref{ex:argus}.

\begin{figure}
	\centering
	\begin{tabular}{|c|ccc|}
		\hline
		& Train & Val. & Test \\
		\hline
		Original $\#q$ & 1829 & 303 & 295 \\
		Post-search $\#q$ & 1081 & 167 & 158 \\
		Average $\#m$ per q. & 19.04 & 13.99 & 16.66 \\
		\hline
	\end{tabular}
	\vspace*{-0.2cm}
	\caption{\footnotesize%
		Characteristics of the Argus QA dataset.
	}
	\label{tab:dataset}
\end{figure}


\begin{figure*}
	\begin{tabular}{|l|}
		\hline
	\textbf{Will Andre Iguodala win NBA Finals MVP in 2015?} \\
		Should Andre Iguodala have won the NBA Finals MVP award over LeBron James? \\
		12.12am ET   Andre Iguodala was named NBA Finals MVP, not LeBron. \\
	\hline
	\textbf{Will Donald Trump run for President in 2016?} \\
		Donald Trump released “Immigration Reform that will make America Great Again” last weekend --- \\ \dots his first, detailed position paper since announcing his campaign for the Republican nomination \\ \dots for president. \\
		The Fix: A brief history of Donald Trump blaming everything on President Obama \\
		DONALD TRUMP FOR PRESIDENT OF PLUTO! \\
	\hline
	\end{tabular}
	\vspace*{-0.2cm}
	\caption{\footnotesize%
		Example pairs in the Argus dataset.
	}
	\label{ex:argus}
\end{figure*}

\begin{figure*}
	\begin{tabular}{|l|}
		\hline
	\textbf{\textit{pedigree chart model} is used to show the pattern of traits that are passed from one generation} \\ \textbf{to the next in a family?} \\
		A pedigree is a chart which shows the inheritance of a trait over several generations. \\
		Figure 51.14 In a pedigree, squares symbolize males, and circles represent females. \\
		\hline
	\textbf{\textit{energy pyramid model} is used to show the pattern of traits that are passed from one generation} \\ \textbf{to the next in a family?} \\
		Energy is passed up a food chain or web from lower to higher trophic levels. \\
		Each step of the food chain in the energy pyramid is called a trophic level. \\
		\hline
	\end{tabular}
	\vspace*{-0.2cm}
	\caption{\footnotesize%
		Example pairs in the AI2-8grade/CK12 dataset.
		Answer texts substituted to a question are shown in italics.
	}
	\label{ex:ai2}
\end{figure*}

\subsection{AI2-8grade/CK12 Dataset}

The \textbf{AI2 Elementary School Science Questions} (no-diagrams variant)%
\footnote{\url{http://allenai.org/data.html}} released by the Allen Institute
cover 855 basic four-choice questions regarding high school science
and follows up to the Allen AI Science Kaggle challenge.%
\footnote{\url{https://www.kaggle.com/c/the-allen-ai-science-challenge}}
The vocabulary includes scientific jargon and named entities, and many questions
are not factoid, requiring real-world reasoning or thought experiments.

We have combined each answer with the respective question (by
substituting the \textit{wh}-word in the question by each answer) and retrieved
evidence sentences for each hypothesis using Solr search in a collection
of CK-12 ``Concepts B'' textbooks.%
\footnote{We have also tried English Wikipedia, but the dataset is much harder.}
525 questions attained any supporting evidence,
examples are shown in Fig.~\ref{ex:ai2}.

We consider this dataset as \textit{preliminary} since it was not reviewed
by a human and many hypotheses are apparently unprovable by the evidence
we have gathered (i.e.\ the theoretical top accuracy is much lower
than 1.0).  However, we released it to the public%
\footnote{\url{https://github.com/brmson/dataset-sts} directory \texttt{data/hypev/ai2-8grade}}
and still included it in the comparison as these
qualities reflect many realistic datasets of unknown qualities, so
we find relative performances of models on such datasets instructive.

\subsection{MCTest Dataset}

The \textbf{Machine Comprehension Test} \cite{MCTest} dataset has been introduced
to provide a challenge for researchers to come up with models that approach
human-level reading comprehension, and serve as a higher-level alternative
to semantic parsing tasks that enforce a specific knowledge representation.
The dataset consists of a set of 660 stories spanning
multiple sentences, written in simple and clean language (but with less restricted
vocabulary than e.g.\ the bAbI dataset \cite{bAbI}).  Each story is accompanied
by four questions and each of these lists four possible answers; the questions are tagged
as based on just \textit{one} in-story sentence, or requiring \textit{multiple} sentence inference.
We use an official extension of the dataset for RTE evaluation
that again textually merges questions and answers.

The dataset is split in two parts, MC-160 and MC-500, based on provenance but
similar in quality.
We train all models on a joined training set.

The practical setting differs from the Argus task as the MCTest dataset contains
relatively restricted vocabulary and well-formed sentences. Furthermore,
the goal is to find the single key point in the story to focus on, while
in the Argus setting we may have many pieces of evidence supporting an answer;
another specific characteristics of MCTest is that it consists of stories
where the ordering and proximity of evidence sentences matters.

\section{Related Work}
\label{sec:relwork}

Our primary concern when integrating natural language query with
textual evidence is to find sentence-level representations suitable
both for relevance weighing and answer prediction.

Sentence-level representations in the retrieval + inference context have been
popularly proposed within the Memory Network framework \cite{MemNN},
but explored just in the form of averaged word embeddings; the task includes
only very simple sentences and a small vocabulary.
Much more realistic setting is introduced in the Answer Sentence Selection
context \cite{AnsselWang} \cite{sps}, with state-of-art models using complex
deep neural architectures with attention \cite{attnpooling}, but the selection
task consists of only retrieval and no inference (answer prediction).
A more indirect retrieval task regarding news summarization was investigated
by \cite{AttSum}.

In the entailment context, \cite{SNLI} introduced a large dataset
with single-evidence sentence pairs (Stanford Natural Language Inference, SNLI),
but a larger vocabulary and
slightly more complicated (but still conservatively formed) sentences.
They also proposed baseline recurrent neural model for modeling
sentence representations, while word-level attention based models
are being studied more recently \cite{SNLIattn} \cite{LSTMMR}.


In the MCTest text comprehension challenge \cite{MCTest}, the
leading models use complex engineered features ensembling multiple traditional
semantic NLP approaches \cite{MCWang}. The best deep model so far
\cite{HABCNN} uses convolutional neural networks for sentence
representations, and attention on multiple levels to pick evidencing
sentences.

\section{Neural Model}
\label{sec:model}

Our approach is to use a sequence of word embeddings to build
sentence embeddings for each hypothesis and respective evidence,
then use the sentence embeddings to estimate relevance and
entailment of each evidence with regard to the respective
hypothesis, and finally integrate the evidence to a single answer.

\subsection{Sentence Embeddings}

To produce sentence embeddings, we investigated the
neural models proposed in the \texttt{da\-ta\-set-sts} framework
for deep learning of sentence pair scoring functions. \cite{sps}

We refer the reader to \cite{sps} and its references for detailed
model descriptions.  We evaluate an \textbf{RNN} model which uses bidirectionally
summed GRU memory cells \cite{GRU} and uses the final states as embeddings;
a \textbf{CNN} model which uses sentence-max-pooled convolutional filters as embeddings \cite{KimMultichannelCNN};
an \textbf{RNN-CNN} model which puts the CNN on top of per-token GRU outputs
rather than the word embeddings \cite{attn1511}; and an \textbf{attn1511} model
inspired by \cite{attn1511} that integrates the RNN-CNN model
with per-word attention to build hypothesis-specific evidence embeddings.
We also report the baseline results of
\textbf{avg} mean of word embeddings in the sentence with projection matrix
and \textbf{DAN} Deep Averaging Network model that employs word-level dropout
and adds multiple nonlinear transformations on top of the averaged embeddings \cite{DAN}.

The original \textbf{attn1511} model \cite{sps} (as tuned for the Answer Sentence Selection task)
used a softmax attention mechanism that would effectively select only a few key words
of the evidence to focus on --- for a hypothesis-evidence token $t$ scalar attention score $a_{h,e}(t)$,
the focus $s_{h,e}(t)$ is:
$$ s_{h,e}(t) = \exp(a_{h,e}(t)) / \sum_{t'} \exp(a_{h,e}(t')) $$
A different focus mechanism exhibited better performance in the Hypothesis Evaluation task,
modelling per-token attention more independently:
$$ s_{h,e}(t) = \sigma(a_{h,e}(t)) / \max_{t'} \sigma(a_{h,e}(t')) $$
We also use relu instead of tanh in the CNNs.

As model input, we use the standard GloVe embeddings \cite{GloVe}
extended with binary inputs denoting token type and overlap with token or bigram in the paired sentence,
as described in \cite{sps}.
However, we introduce two changes to the word
embedding model --- we use 50-dimensional embeddings instead of 300-dimensional,
and rather than building an adaptable embedding matrix from the
training set words preinitialized by GloVe, we use only the top 100
most frequent tokens in the adaptable embedding matrix and use
fixed GloVe vectors for all other tokens (including tokens not
found in the training set).  In preliminary experiments, this improved generalization
for highly vocabulary-rich tasks like Argus, while still allowing the high-frequency
tokens (like interpunction or conjunctions) to learn semantic operator representations.

As an additional method for producing sentence embeddings, we consider
the \textbf{Ubu.\ RNN} transfer learning method proposed by \cite{sps} where an RNN model
(as described above) is trained on the Ubuntu Dialogue task \cite{UbuntuLowe}.%
\footnote{The Ubuntu Dialogue dataset consists of one million chat dialog contexts,
learning to rank candidates for the next utterance in the dialog;
the sentences are based on IRC chat logs of the Ubuntu
community technical support channels and contain casually typed interactions regarding
computer-related problems, resembling tweet data, but longer and with heavily technical jargon.}
The pretrained model weights are used to initialize an RNN model which is then fine-tuned
on the Hypothesis Evaluation task.  We use the same model as originally proposed (except the aforementioned vocabulary handling modification),
with the dot-product scoring used for Ubuntu Dialogue training replaced by MLP point-scores described below.

\subsection{Evidence Integration}

Our main proposed schema for evidence integration is \textbf{Evidence Weighing}.
From each pair of hypothesis and evidence embeddings,%
\footnote{We employ Siamese training, sharing the weights between hypothesis and evidence embedding models.}
we produce two $[0,1]$
predictions using a pair of MLP point-scorers of \texttt{dataset-sts} \cite{sps}%
\footnote{From the elementwise product
and sum of the embeddings, a linear classifier directly produces a prediction;
contrary to the typical setup, we use no hidden layer.}
with sigmoid activation function.  The predictions are interpreted as $C_i \in [0,1]$
entailment (0 to 1 as \textit{no} to \textit{yes}) and relevance $R_i \in [0,1]$.
To integrate the predictions across multiple pieces of evidence,
we propose a weighed average model:

$$ y  = \frac{\sum_i C_iR_i}{\sum_i R_i} $$

We do not have access to any explicit labels for the evidence,
but we train the model end-to-end with just $y$ labels and the formula
for $y$ is differentiable, carrying over the gradient to the sentence
embedding model.
This can be thought of as a simple passage-wide attention model.

As a baseline strategy, we also consider \textbf{Evidence Averaging}, where
we simply produce a single scalar prediction per hypothesis-evidence pair
(using the same strategy as above) and decide the hypothesis simply based
on the mean prediction across available evidence.

Finally, following success reported in the Answer Sentence Selection task \cite{sps},
we consider a \textbf{BM25 Feature} combined with Evidence Averaging,
where the MLP scorer that produces the pair scalar prediction as above takes
an additional BM25 word overlap score input \cite{BM25} besides the elementwise
embedding comparisons.

\section{Results}
\label{sec:res}

\subsection{Experimental Setup}

We implement the differentiable model in the Keras framework \cite{Keras} and
train the whole network from word embeddings to output evidence-integrated
hypothesis label using the binary cross-entropy loss as an objective%
\footnote{Unlike \cite{HABCNN}, we have found ranking-based loss functions
	ineffective for this task.}
and the Adam optimization algorithm \cite{Adam}.  We apply $\mathbb{L}_2 = 10^{-4}$
regularization and a $p=1/3$ dropout.

Following the recommendation of \cite{sps}, we report expected test set
question accuracy%
\footnote{In the MCTest and AI2-8grade/CK12 datasets, we test and rank four
	hypotheses per question, whereas in the Argus dataset, each
	hypothesis is a single question.}
as determined by average accuracy in 16 independent trainings
and with 95\% confidence intervals based on the Student's
t-distribution.

\subsection{Evaluation}

\begin{figure}[t]
\centering
\setlength{\tabcolsep}{3pt}
\begin{tabular}{|c|c|c|c|}
\hline
Model              & train & val & test \\
\hline
avg & $0.872$ & $0.816$ & $0.744$\\
 & $\quad^{\pm0.009}$ & $\quad^{\pm0.008}$ & $\quad^{\pm0.020}$\\
DAN & $0.884$ & $0.822$ & $0.754$\\
 & $\quad^{\pm0.012}$ & $\quad^{\pm0.011}$ & $\quad^{\pm0.025}$\\
\hline
RNN & $0.906$ & $0.875$ & $0.823$\\
 & $\quad^{\pm0.013}$ & $\quad^{\pm0.005}$ & $\quad^{\pm0.008}$\\
CNN & $0.896$ & $0.857$ & $0.822$\\
 & $\quad^{\pm0.018}$ & $\quad^{\pm0.006}$ & $\quad^{\pm0.007}$\\
RNN-CNN & $0.885$ & $0.860$ & $0.816$\\
 & $\quad^{\pm0.010}$ & $\quad^{\pm0.007}$ & $\quad^{\pm0.009}$\\
attn1511 & $0.935$ & $0.877$ & $0.816$\\
 & $\quad^{\pm0.021}$ & $\quad^{\pm0.008}$ & $\quad^{\pm0.008}$\\
Ubu.\ RNN & $0.951$ & $0.912$ & \textbf{0.852}\\
 & $\quad^{\pm0.017}$ & $\quad^{\pm0.004}$ & $\quad^{\pm0.008}$\\
\hline
\end{tabular}
\setlength{\tabcolsep}{6pt}
\vspace*{-0.2cm}
\caption{\footnotesize%
	Model accuracy on the Argus task, using the evidence weighing scheme.
}
\label{tab:argus}
\end{figure}

\begin{figure}[t]
\centering
\setlength{\tabcolsep}{3pt}
\begin{tabular}{|c|c|c|c|}
\hline
Model              & Mean Ev. & BM25 Feat. & Weighed \\
\hline
avg & $0.746$ & $0.770$ & $0.744$\\
 & $\quad^{\pm0.051}$ & $\quad^{\pm0.011}$ & $\quad^{\pm0.020}$\\
\hline
RNN & $0.822$ & $0.828$ & $0.823$\\
 & $\quad^{\pm0.015}$ & $\quad^{\pm0.015}$ & $\quad^{\pm0.008}$\\
attn1511 & $0.819$ & $0.811$ & $0.816$\\
 & $\quad^{\pm0.013}$ & $\quad^{\pm0.012}$ & $\quad^{\pm0.008}$\\
Ubu.\ RNN & $0.847$ & $0.831$ & $0.852$\\
 & $\quad^{\pm0.009}$ & $\quad^{\pm0.018}$ & $\quad^{\pm0.008}$\\
\hline
\end{tabular}
\setlength{\tabcolsep}{6pt}
\vspace*{-0.2cm}
\caption{\footnotesize%
	Comparing the influence of the evidence integration schemes on the Argus test accuracy.
}
\label{tab:argusev}
\end{figure}

In Fig.~\ref{tab:argus}, we report the model performance on the Argus task,
showing that the Ubuntu Dialogue transfer RNN outperforms other proposed
models by a large margin.  However, a comparison of evidence integration
approaches in Fig.~\ref{tab:argusev} shows that
evidence integration is not the major deciding factor and there are no
staticially meaningful differences between the evaluated approaches.
We measured high correlation between classification and relevance scores
with Pearson's $r = 0.803$, showing that our model does not learn a separate
evidence weighing function on this task.

\begin{figure}[t]
\centering
\setlength{\tabcolsep}{3pt}
\begin{tabular}{|c|c|c|c|}
\hline
Model              & train & val & test \\
\hline
avg & $0.505$ & $0.442$ & $0.401$\\
 & $\quad^{\pm0.024}$ & $\quad^{\pm0.022}$ & $\quad^{\pm0.016}$\\
DAN & $0.556$ & $0.491$ & $0.391$\\
 & $\quad^{\pm0.038}$ & $\quad^{\pm0.015}$ & $\quad^{\pm0.008}$\\
\hline
RNN & $0.712$ & $0.381$ & $0.361$\\
 & $\quad^{\pm0.053}$ & $\quad^{\pm0.016}$ & $\quad^{\pm0.012}$\\
CNN & $0.676$ & $0.442$ & $0.384$\\
 & $\quad^{\pm0.056}$ & $\quad^{\pm0.012}$ & $\quad^{\pm0.011}$\\
RNN-CNN & $0.582$ & $0.439$ & $0.376$\\
 & $\quad^{\pm0.057}$ & $\quad^{\pm0.024}$ & $\quad^{\pm0.014}$\\
attn1511 & $0.725$ & $0.384$ & $0.358$\\
 & $\quad^{\pm0.069}$ & $\quad^{\pm0.012}$ & $\quad^{\pm0.015}$\\
Ubu.\ RNN & $0.570$ & $0.494$ & \textbf{0.441}\\
 & $\quad^{\pm0.059}$ & $\quad^{\pm0.012}$ & $\quad^{\pm0.011}$\\
\hline
\end{tabular}
\setlength{\tabcolsep}{6pt}
\vspace*{-0.2cm}
\caption{\footnotesize%
	Model (question-level) accuracy on the AI2-8grade/CK12 task, using the evidence weighing scheme.
}
\label{tab:ai2}
\end{figure}

\begin{figure}[t]
\centering
\setlength{\tabcolsep}{3pt}
\begin{tabular}{|c|c|c|c|}
\hline
Model              & Mean Ev. & BM25 Feat. & Weighed \\
\hline
avg & $0.366$ & \textbf{0.415} & \textbf{0.401}\\
 & $\quad^{\pm0.010}$ & $\quad^{\pm0.008}$ & $\quad^{\pm0.016}$\\
\hline
CNN & $0.385$ &  & $0.384$\\
 & $\quad^{\pm0.020}$ &  & $\quad^{\pm0.011}$\\
Ubu.\ RNN & $0.416$ & $0.418$ & \textbf{0.441}\\
 & $\quad^{\pm0.011}$ & $\quad^{\pm0.009}$ & $\quad^{\pm0.011}$\\
\hline
\end{tabular}
\setlength{\tabcolsep}{6pt}
\vspace*{-0.2cm}
\caption{\footnotesize%
	Comparing the influence of the evidence integration schemes on the AI2-8grade/CK12 test accuracy.
}
\label{tab:ai2ev}
\end{figure}

In Fig.~\ref{tab:ai2}, we look at the model performance on the AI2-8grade/CK12 task,
repeating the story of Ubuntu Dialogue transfer RNN dominating other models.
However, on this task our proposed evidence weighing scheme improves over
simpler approaches --- but just on the best model, as shown in Fig.~\ref{tab:ai2ev}.
On the other hand, the simplest averaging model benefits from at least BM25
information to select relevant evidence, apparently.

\begin{figure*}[t]
\centering
\setlength{\tabcolsep}{3pt}
\begin{tabular}{|c|c|ccc|ccc|}
\hline
& joint & MC-160     &             &            & MC-500     &      & \\
Model       & all (train)        & one & multi & all & one & multi & all \\
\hline
\hline
hand-crafted &  & $0.842$ & $0.678$ & $0.753$ & $0.721$ & $0.679$ & $0.699$\\
\hline
\hline
Attn. Reader & & $0.481$ & $0.447$ & $0.463$ & $0.444$ & $0.395$ & $0.419$ \\
Neur. Reasoner & & $0.484$ & $0.468$ & $0.476$ & $0.457$ & $0.456$ & $0.456$ \\
HABCNN-TE & & $0.633$ & \textbf{0.629} & \textbf{0.631} & $0.542$ & \textbf{0.517} & $0.529$ \\
\hline
\hline
avg & $0.577$ & $0.653$ & $0.471$ & $0.556$ & $0.587$ & $0.506$ & \textbf{0.542}\\
 & $\quad^{\pm0.009}$ & $\quad^{\pm0.027}$ & $\quad^{\pm0.020}$ & $\quad^{\pm0.012}$ & $\quad^{\pm0.018}$ & $\quad^{\pm0.010}$ & $\quad^{\pm0.011}$\\
DAN & $0.590$ & $0.681$ & $0.486$ & $0.577$ & \textbf{0.636} & $0.496$ & \textbf{0.560}\\
 & $\quad^{\pm0.009}$ & $\quad^{\pm0.017}$ & $\quad^{\pm0.010}$ & $\quad^{\pm0.010}$ & $\quad^{\pm0.013}$ & $\quad^{\pm0.007}$ & $\quad^{\pm0.007}$\\
\hline
RNN & $0.608$ & $0.583$ & $0.490$ & $0.533$ & $0.539$ & $0.456$ & $0.494$\\
 & $\quad^{\pm0.030}$ & $\quad^{\pm0.033}$ & $\quad^{\pm0.018}$ & $\quad^{\pm0.020}$ & $\quad^{\pm0.016}$ & $\quad^{\pm0.013}$ & $\quad^{\pm0.012}$\\
CNN & $0.658$ & $0.655$ & $0.511$ & $0.578$ & $0.571$ & $0.483$ & $0.522$\\
 & $\quad^{\pm0.021}$ & $\quad^{\pm0.020}$ & $\quad^{\pm0.012}$ & $\quad^{\pm0.014}$ & $\quad^{\pm0.013}$ & $\quad^{\pm0.012}$ & $\quad^{\pm0.009}$\\
RNN-CNN & $0.597$ & $0.617$ & $0.493$ & $0.551$ & $0.554$ & $0.470$ & $0.508$\\
 & $\quad^{\pm0.039}$ & $\quad^{\pm0.041}$ & $\quad^{\pm0.021}$ & $\quad^{\pm0.020}$ & $\quad^{\pm0.023}$ & $\quad^{\pm0.016}$ & $\quad^{\pm0.014}$\\
attn1511 & $0.687$ & $0.611$ & $0.485$ & $0.544$ & $0.571$ & $0.454$ & $0.507$\\
 & $\quad^{\pm0.061}$ & $\quad^{\pm0.052}$ & $\quad^{\pm0.025}$ & $\quad^{\pm0.033}$ & $\quad^{\pm0.036}$ & $\quad^{\pm0.011}$ & $\quad^{\pm0.021}$\\
Ubu.\ RNN & $0.678$ & \textbf{0.736} & $0.503$ & \textbf{0.612} & \textbf{0.641} & $0.452$ & \textbf{0.538}\\
 & $\quad^{\pm0.035}$ & $\quad^{\pm0.033}$ & $\quad^{\pm0.016}$ & $\quad^{\pm0.023}$ & $\quad^{\pm0.017}$ & $\quad^{\pm0.017}$ & $\quad^{\pm0.015}$\\
\hline
$^*$ Ubu.\ RNN & & $0.786$ & $0.547$ & $0.658$ & $0.676$ & $0.494$ & $0.577$ \\
\hline
\end{tabular}
\setlength{\tabcolsep}{6pt}
\vspace*{-0.2cm}
\caption{\footnotesize%
	Model (question-level) accuracy on the test split of the MCTest task, using the evidence weighing scheme.
	The first column shows accuracy on a train split joined across both datasets.\\
	$^*$ The model with top MC-500 test set result (across 16 runs)
	that convincingly dominates HABCNN-TE in the \textit{one} and \textit{all} classes and illustrates that the issue of
	reporting evaluation spread is not just theoretical. 5/16 of the models have MC-160 \textit{all} accuracy $>0.631$.
}
\label{tab:mctest}
\end{figure*}

\begin{figure}[t]
\centering
\setlength{\tabcolsep}{3pt}
\begin{tabular}{|c|c|c|c|}
\hline
Model              & Mean Ev. & BM25 Feat. & Weighed \\
\hline
avg & $0.423$ & $0.506$ & \textbf{0.542}\\
 & $\quad^{\pm0.014}$ & $\quad^{\pm0.012}$ & $\quad^{\pm0.011}$\\
\hline
CNN & $0.373$ & \textbf{0.509} & \textbf{0.522}\\
 & $\quad^{\pm0.036}$ & $\quad^{\pm0.027}$ & $\quad^{\pm0.009}$\\
Ubu.\ RNN & $0.507$ & $0.509$ & \textbf{0.538}\\
 & $\quad^{\pm0.014}$ & $\quad^{\pm0.012}$ & $\quad^{\pm0.015}$\\
\hline
\end{tabular}
\setlength{\tabcolsep}{6pt}
\vspace*{-0.2cm}
\caption{\footnotesize%
	Comparing the influence of the evidence integration schemes on the MC-500 (all-type) test accuracy.
}
\label{tab:mctestev}
\end{figure}

For the MCTest dataset,
Fig.~\ref{tab:mctest} compares our proposed models with the current state-of-art
ensemble of hand-crafted syntactic and frame-semantic features \cite{MCWang},
as well as past neural models
from the literature, all using attention mechanisms --- the Attentive Reader of \cite{ReadComprehend},
Neural Reasoner of \cite{NeuralReasoner} and the HABCNN model family of \cite{HABCNN}.%
\footnote{\cite{HABCNN} also reports the results on the former models.}
We see that averaging-based models are surprisingly effective on this task,
and in particular on the MC-500 dataset it can beat even the best so far reported model of HABCNN-TE.\@
Our proposed transfer model is statistically equivalent to the best model on both datasets
(furthermore, previous work did not include
confidence intervals, even though their models should also be stochastically initialized).

As expected, our models did badly on the multiple-evidence class of questions
--- we made no attempt to model information flow across adjacent sentences in our
models as this aspect is unique to MCTest in the
context of our work.


Interestingly, evidence weighing does play an important role on the MCTest task
as shown in Fig.~\ref{tab:mctestev},
significantly boosting model accuracy.
This confirms that a mechanism to allocate attention to different sentences
is indeed crucial for this task.

\subsection{Analysis}

While we can universally proclaim Ubu.\ RNN as the best model,
we observe many aspects of the Hypothesis Evaluation problem that are shared
by the AI2-8grade/CK12 and MCTest tasks, but not by the Argus task.

Our largest surprise lies in the ineffectivity of evidence weighing
on the Argus task, since observations of irrelevant passages initially
led us to investigate this model.  
We may also see that non-pretrained RNN does very well on the Argus task
while CNN is a better model otherwise.

An aspect that could explain this rift is that the latter two tasks
are primarily \textit{retrieval} based, where we seek to judge each
evidence as irrelevant or essentially a paraphrase of the hypothesis.
On the other hand, the Argus task is highly \textit{semantic}
and compositional, with the questions often differing just by a presence
of negation --- recurrent model that can capture long-term dependencies
and alter sentence representations based on the presence of negation
may represent an essential improvement over an n-gram-like convolutional
scheme.
We might also attribute the lack of success of evidence weighing in the Argus task
to a more conservative scheme of passage retrieval employed
in the IR pipeline that produced the dataset.
Given the large vocabulary and noise levels in the data, we may also simply
require more data to train the evidence weighing properly.

We see from the training vs.\ test accuracies that RNN-based
models (including the word-level attention model) have a strong tendency
to overfit on our small datasets, while CNN is much more resilient.
While word-level attention seems appealing for such a task,
we speculate that we simply might not have enough training data to
properly train it.%
\footnote{Just reducing the dimensionality of hidden representations did not yield an improvement.}
Investigating attention transfer is a point for future work --- by our
preliminary experiments on multiple datasets, attention
models appear more task specific than the basic text comprehension models
of memory based RNNs.


One concrete limitation of our models in case of the Argus task
is a problem of reconciling particular named entity instances.
The more obvious form of this issue is \textit{Had Roger Federer beat Martin Cilic in US OPEN 2014?}
versus an opposite \textit{Had Martin Cilic beat Roger Federer in US OPEN 2014?} ---
another form of this problem is reconciling a hypothesis like
\textit{Will the Royals win the World Series?}
with evidence
\textit{Giants Win World Series With Game 7 Victory Over Royals}.
An abstract embedding of the sentence will not carry over the required
information --- it is important to explicitly pass and reconcile the roles
of multiple named entities which cannot be meaningfully embedded in a GloVe-like
semantic vector space.

\section{Conclusion}
\label{sec:concl}

We have established a general Hypothesis Evaluation task with three datasets
of various properties, and shown that neural models can exhibit strong
performance (with less hand-crafting effort than non-neural classifiers).
We propose an evidence weighing model that is never harmful and improves
performance on some tasks.
We also demonstrate that simple models can outperform or closely match
performance of complex architectures;
all the models we consider are task-independent and were successfully
used in different contexts than Hypothesis Evaluation \cite{sps}.
Our results empirically show that a basic RNN text comprehension model
well trained on a large dataset (even if the task is unrelated and vocabulary
characteristics are very different) outperforms or matches more complex
architectures trained only on the dataset of the task at hand.%
\footnote{Even if these use multi-task learning, which was employed in case
of the HABCNN models that were trained to also predict question classes.}

Finally, on the MCTest dataset, our best proposed model is better
or statistically
indistinguishable from the best neural model reported so far
\cite{HABCNN},
even though it has a simpler architecture and only a naive attention
mechanism.

We would like to draw several recommendations for future research from our findings:
(A) encourage usage of basic neural architectures as evaluation baselines;
(B) suggest that future research includes models pretrained on large data as baselines;
(C) validate complex architectures on tasks with large datasets if they cannot beat baselines on small datasets; and
(D) for randomized machine comprehension models (e.g.\ neural networks with random weight
initialization, batch shuffling or probabilistic dropout), report expected test set performance
based on multiple independent training runs.

As a general advice for solving complex tasks with small datasets,
besides the point (B) above our analysis suggests convolutional networks
as the best models regarding the tendency to overfit, unless
semantic composionality plays a crucial role in the task;
in this scenario, simple averaging-based models are a great start as well.
Preinitializing a model also helps against overfitting.

We release our implementation of the Argus task, evidence integration
models and processing of all the evaluated datasets as open source.%
\footnote{\url{https://github.com/brmson/dataset-sts} task \textit{hypev}}

We believe the next step towards machine comprehension NLP models
(based on deep learning but capable of dealing with real-world,
large-vocabulary data) will involve research into a better way to deal with
entities without available embeddings. 
When distinguishing specific entities,
simple word-level attention mechanisms will not do.
A promising approach could extend
the flexibility of the final sentence representation, moving from attention mechanism
to a memory mechanism\footnote{Not necessarily ``memories'' in the sense of Memory Networks.}
by allowing the network to remember a set of ``facts'' derived from each sentence;
related work has been done for example on end-to-end differentiable shift-reduce parsers
with LSTM as stack cells \cite{EndToEndParsing}.

\section*{Acknowledgments}
{\footnotesize
	This work was co-funded by the Augur Project of the Forecast Foundation
and financially supported by the Grant Agency of the Czech Technical
University in Prague, grant No. SGS16/ 084/OHK3/1T/13.
Computational resources were provided by the CESNET LM2015042 and the CERIT Scientific Cloud LM2015085,
provided under the programme ``Projects of Large Research, Development, and Innovations Infrastructures.''

We'd like to thank Peronet Despeignes of the Augur Project for his support.
Carl Burke has provided instructions for searching CK-12 ebooks within
the Kaggle challenge.}

\bibliography{qa,sps,argus}
\bibliographystyle{acl2016}

\end{document}